%% file: main.tex
\documentclass{MITcsail}

\input{math_commands}

\input{mlVecMat}

\usepackage{listings}
\usepackage{float}
\usepackage{graphicx}
\usepackage{placeins}

\newcommand{\method}{\textsc{JEPA-TTT}}
\newcommand{\pagehead}{\method}

\title{\method: Persistent Test-Time Training of Latent World Models for Planning under Dynamics Shifts}

\author[1]
{Zheyuan Zhang\textsuperscript{1,2,$\dagger$}, Suyu Ye\textsuperscript{1,2,$\dagger$}, Nakul Agarwal\textsuperscript{1}, Hossein Nourkhiz Mahjoub\textsuperscript{1}, \\ \bf \sffamily Ehsan Moradi Pari\textsuperscript{1}, Daniel Khashabi\textsuperscript{2}, Tianmin Shu\textsuperscript{2}, Vaishnav Tadiparthi\textsuperscript{1} \\
\vspace{0.6em}
{\normalfont\sffamily \textsuperscript{1}Honda Research Institute USA \quad \textsuperscript{2}Johns Hopkins University} \\
\vspace{0.6em}
\texttt{\href{https://jepa-ttt.github.io/}{jepa-ttt.github.io}}
}

\begin{document}

\maketitle
\footnotetext[2]{Work done during an internship at Honda Research Institute USA}

\thispagestyle{firstpagestyle}

\paragraph{\textit{Abstract.}}
World models enable agents to plan by predicting future states of the environment, but their predictions can become unreliable when test-time dynamics differ from those seen during training. We present \method{}, which adapts the latent dynamics predictor of a pretrained action-conditioned Joint-Embedding Predictive Architecture world model throughout test time. Self-supervised updates accumulate across episodes, while the visual encoder and reward head remain fixed, preserving the pretrained representation and task objective. Planning requires neither a goal image nor online environment reward. \method{} uses dense replay, which forms prediction windows at every temporal offset, retains them in a growing buffer, and samples minibatches from that buffer for predictor updates. Across eight dynamics shifts in four continuous-control environments, \method{} improves planning on every shift. After 500 test-time episodes, it reduces autoregressive latent prediction error by 83\% on average and improves planning performance by 153\% over the frozen JEPA world model. These results show that persistent self-supervised test-time training can adapt a pretrained latent world model under changed dynamics.

\input{sections/introduction}
\input{sections/related_work}
\input{sections/method}
\input{sections/experiments}
\input{sections/conclusion}

\newpage

\bibliography{references}

\input{sections/appendix}

\end{document}

%% file: math_commands.tex
\usepackage{amsmath,amsfonts,bm}

\def\eqref#1{equation~\ref{#1}}

\def\1{\bm{1}}

\DeclareMathAlphabet{\mathsfit}{\encodingdefault}{\sfdefault}{m}{sl}
\SetMathAlphabet{\mathsfit}{bold}{\encodingdefault}{\sfdefault}{bx}{n}



%% file: mlVecMat.tex
\definecolor{darkred}{RGB}{140, 21, 21}
\definecolor{lightgray}{gray}{0.5}
\definecolor{orange}{HTML}{F58025}
\definecolor{deepred}{rgb}{0.631,0.102,0.102}
\definecolor{amethyst}{rgb}{0.6, 0.4, 0.8}
\definecolor{darkgreen}{rgb}{0.3,0.7,0.3}
\definecolor{salmon}{RGB}{241, 150, 141}
\definecolor{mildyellow}{HTML}{FFF2CC}
\definecolor{jhulightblue}{HTML}{68ACE5}

\usepackage{hyperref}
\usepackage{xcolor}
\hypersetup{
    colorlinks=true,
    linkcolor=jhulightblue,
    citecolor=lightgray,
    filecolor=darkred,
    urlcolor=jhulightblue
}
\usepackage[authoryear]{natbib}
\setcitestyle{square,aysep={,}}
\usepackage{amsmath}
\usepackage{amsfonts}
\usepackage{amssymb}
\usepackage{wrapfig}
\usepackage{subcaption}
\usepackage{multirow}
 \usepackage{mathtools} 

\usepackage{verbatim}

\usepackage{anyfontsize}

\usepackage{microtype}
\usepackage{graphicx}
\usepackage{booktabs} 
\usepackage{multirow}
\usepackage{amsmath,amssymb}
\usepackage{booktabs}
\usepackage{caption,subcaption}
\usepackage{svg}

\usepackage{authblk}

\renewcommand\Affilfont{\normalsize\normalfont}

\definecolor{mygreen}{HTML}{3cb44b}
\definecolor{skyblue}{HTML}{beffff}
\definecolor{lightgreen}{HTML}{90ee90}

\usepackage{color, colortbl}

\definecolor{emerald}{rgb}{0.31, 0.78, 0.37}

\usepackage{datetime}
\newdateformat{ymd}{\the\year-\the\month-\the\day}

\usepackage{tcolorbox}
\usepackage{enumitem}
\definecolor{mygreen}{HTML}{3cb44b}
\colorlet{myyellow}{green!10!orange!90!}
\makeatletter

\usepackage{tikz}
\usetikzlibrary{arrows,shapes,snakes,automata,backgrounds,fit,petri}
\usepackage{adjustbox}

\newcommand{\RN}[1]{%
	\textup{\lowercase\expandafter{\it \romannumeral#1}}%
}
\usepackage{tabu}

\newcommand{\beq}{\vspace{0mm}\begin{equation}}
\newcommand{\eeq}{\vspace{0mm}\end{equation}}
\newcommand{\beqs}{\vspace{0mm}\begin{eqnarray}}
\newcommand{\eeqs}{\vspace{0mm}\end{eqnarray}}
\newcommand{\barr}{\begin{array}}
\newcommand{\earr}{\end{array}}

\usepackage{color, colortbl}
\definecolor{Gray}{gray}{0.93}

\usepackage{lipsum}

\usepackage{pifont}

\usepackage{makecell}

\usepackage{xcolor,amsmath}
\usepackage[ruled,vlined,linesnumbered]{algorithm2e}
\SetKwComment{Comment}{$\triangleright$ }{}
\SetKwInput{KwInput}{Input}
\SetKwInput{KwOutput}{Output}
\DontPrintSemicolon

\usepackage{xcolor}
\definecolor{mygreen}{HTML}{3cb44b}

\SetKwComment{Comment}{\color{green!50!black}\# }{}

\SetKwProg{Function}{def}{:}{}

\SetKwProg{For}{for}{:}{}
\SetKwProg{If}{if}{:}{}

%% file: sections/introduction.tex
\section{Introduction}
\label{sec:introduction}

Agents can learn predictive models of environment dynamics (i.e., world models) to support decision-making \citep{ha2018world}. Action-conditioned Joint-Embedding Predictive Architecture (JEPA) world models predict future latent representations from observations and actions \citep{lecun2022path, maes2026leworldmodel}. The effectiveness of a world model for planning depends critically on the accuracy of its learned dynamics. Conventional world models typically assume that the transition dynamics are the same during training and testing \citep{hafner2019learning,maes2026leworldmodel}.
However, in practice, this assumption can be violated by changes in actuator responses, contact dynamics, or object motion, causing a model trained under one set of dynamics to produce systematic prediction errors at test time and thus leading to poor planning performance.

\begin{figure}[t]
    \centering
    \includegraphics[width=\textwidth]{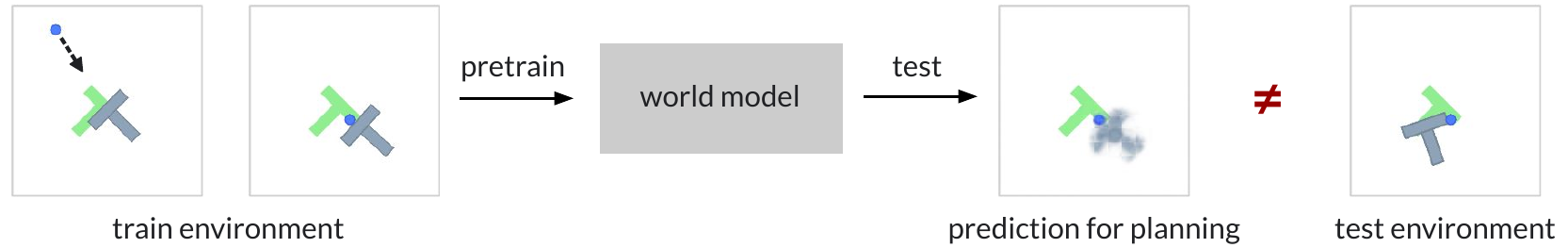}
    \caption{A dynamics shift can cause a pretrained world model to fail during planning. The model predicts the outcome of a candidate action based on the dynamics learned during training, but the test environment has different transitions. This mismatch can cause the planner to select wrong actions.}
    \label{fig:dynamics_shift_problem}
\end{figure}

Figure~\ref{fig:dynamics_shift_problem} illustrates this failure: a pretrained world model predicts the training dynamics, while the action produces a different outcome after the dynamics shifts in the test environment. Given the current observation $o_t$, the agent executes action $a_t$ and receives the next observation $o_{t+1}$. The transition $(o_t,a_t,o_{t+1})$ provides a self-supervised target for updating the dynamics model.

\FloatBarrier

We study whether a pretrained action-conditioned JEPA world model can adapt to new dynamics throughout test time and thereby improve planning. We introduce \method{}, which adapts the model's latent dynamics predictor through self-supervised updates that accumulate across episodes. Since JEPA learns predictive representations without a reconstruction objective, it eliminates the computational cost of a decoder, making it a practical choice for test-time training under a limited compute budget.

The model's visual encoder maps observations to latent states, and its predictor predicts how actions change those states. \method{} updates only the predictor using executed actions and consecutive encoded observations. The visual encoder and offline-trained reward head remain fixed, preserving the visual representation and task objective. During planning, model predictive control (MPC) uses the adapted predictor and fixed reward head to rank candidate actions without requiring goal images or online environment rewards. To train from online experience, \method{} uses dense replay, which forms prediction windows at every temporal offset and samples minibatches from a growing buffer.

\begin{figure}[t]
    \centering
    \includegraphics[width=\textwidth]{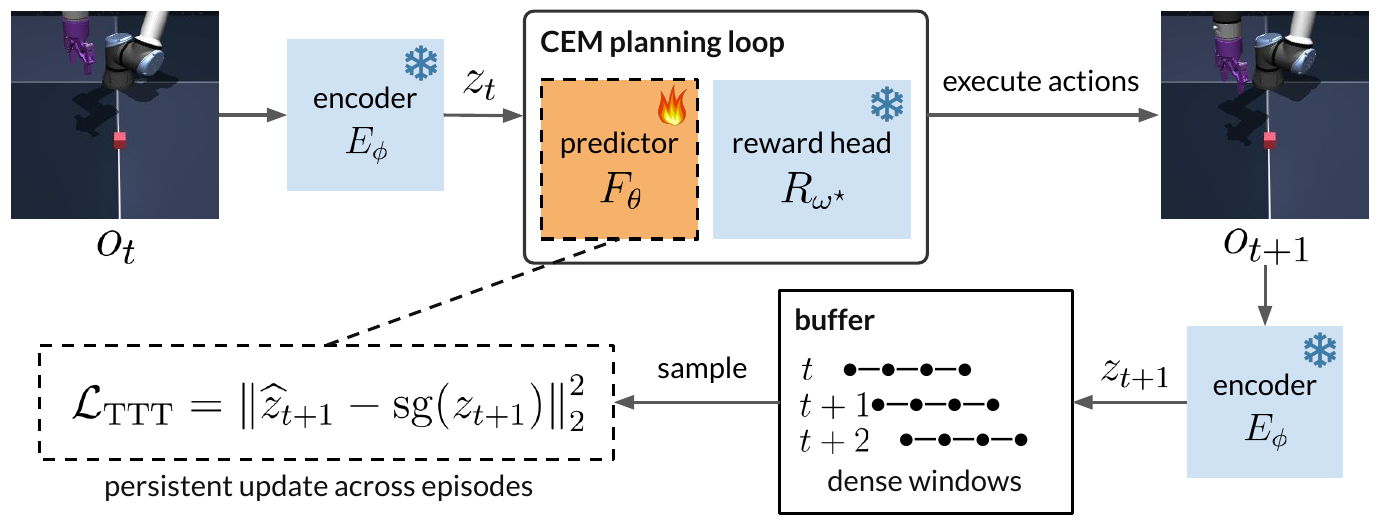}
    \caption{Overview of \method. Cross Entropy Method (CEM) plans with the current latent dynamics predictor $F_\theta$ and frozen reward head $R_{\omega^\star}$. After executing the actions, the frozen visual encoder maps the new observation $o_{t+1}$ to a target latent $z_{t+1}$. \method{} stores prediction windows at every temporal offset and samples from the growing buffer to update $F_\theta$ with a self-supervised latent prediction objective. The adapted predictor is retained across episodes.}
    \label{fig:overview}
\end{figure}

We evaluate two controlled dynamics shifts in each of four continuous-control environments. The shifts span changes to actuation, contact dynamics, and state-dependent action mappings.
\method{} improves planning on all eight shifts. It achieves a mean normalized held-out planning-score AUC of 0.571, compared with 0.267 for the frozen JEPA world model, and raises mean best held-out score from 0.267 to 0.678. After 500 test-time episodes, the adapted predictor reduces autoregressive latent prediction error by 83\% on average. We also conduct ablation studies to assess persistence, temporal coverage, and replay.

\paragraph{Contributions.} To summarize, our main contributions are: (1) \method{}, which adapts a pretrained JEPA predictor for planning without goal images or online environment rewards, using an offline-trained reward head; (2) consistent improvements in planning and latent prediction across eight dynamics shifts; and (3) controlled ablations of test-time training.

%% file: sections/related_work.tex
\section{Related Work}
\label{sec:related_work}

\paragraph{Latent World Models and Joint-Embedding Predictive Architectures.} Generative world models perform prediction in observation space or through explicit scene representations \citep{finn2017deep, du2023learning, du2024video, bar2025navigation, zhang2025combo, zhen2025tesseract, hafner2025training, zhang2026world, yin20263d}, while latent world models predict compact representations that can discard visual details irrelevant to control \citep{ha2018world, hafner2019learning, hafner2019dream}. Dreamer learns behavior through latent imagination, while TD-MPC2 uses latent dynamics for trajectory optimization \citep{hafner2019dream, hansen2024td}. Joint-Embedding Predictive Architectures provide a particularly natural foundation for latent world modeling because they learn predictive representations without requiring pixel reconstruction \citep{lecun2022path, assran2023self, bardes2024revisiting}. Recent works have developed action-conditioned JEPA world models specifically for planning. V-JEPA 2-AC post-trains an action-conditioned predictor for robot planning with image goals \citep{assran2025v}. DINO-WM predicts over frozen pretrained visual features and demonstrates zero-shot latent planning \citep{zhou2024dino}. PLDM learns a latent dynamics model with VICReg-style regularization from offline data \citep{sobal2025learning}. LeJEPA introduces SIGReg for stable end-to-end JEPA training \citep{balestriero2025lejepa}, and LeWorldModel extends this approach to action-conditioned world modeling directly from pixels \citep{maes2026leworldmodel}. \citet{wang2026temporal} reshape latent trajectories to improve their suitability for optimization-based planning. These works primarily study world-model training and planning rather than persistent adaptation to a new dynamics regime. \method{} instead investigates how a pretrained JEPA world model adapts its dynamics online for dynamics shift during deployment.

\paragraph{Test-Time Training and Online Dynamics Adaptation.} Test-time training adapts model parameters using data available during inference, typically through self-supervised objectives or entropy minimization \citep{sun2020test, wang2020tent}. Continual and online variants carry adaptation state across sequential image and video inputs, allowing recent observations to inform subsequent predictions \citep{wang2022continual, wang2025test}. Self-supervised policy adaptation updates visual representations without online rewards \citep{hansen2020self}. \method{} applies this principle to action-conditioned latent dynamics by using observed transitions as self-supervised targets for test-time training. Related work has also viewed parameter updates as a form of memory for long input streams \citep{hardt2024test, sun2024learning}. Online adaptation has a long history in learning-based control, including adaptive dynamics models for MPC \citep{nagabandi2018learning, williams2020locally} and latent environment representations inferred from recent interaction histories \citep{kumar2021rma}. Offline world models can also be finetuned through online model-based RL with environment rewards \citep{feng2023finetuning}. Off-dynamics RL instead corrects the rewards used for source-domain policy training to account for differences in transition dynamics \citep{eysenbach2020off}. \cite{feng2026wam} further explores test-time adaptation from human demonstrations in world-action models. Unlike these approaches, \method{} updates a pretrained JEPA world model directly from the agent's actions and observations in online transitions. The most closely related work is the concurrent work AdaJEPA, which also adapts a JEPA world model from observed transitions within an MPC loop \citep{wang2026adajepa}. AdaJEPA resets its adapted model and buffer at each episode, conditions planning on a goal image that is unavailable in our setting, and updates from a short within-episode history. In contrast, \method{} retains adaptation across episodes, plans with a frozen reward head and no goal image, and uses dense replay over accumulated transitions.

%% file: sections/method.tex

\section{Persistent Test-Time Training of Latent World Models}
\label{sec:method}

\subsection{Problem Setting}\label{sec:problem}

We study finite-horizon control after a single change in transition dynamics that remains fixed throughout test time. The task itself does not change: the state space, action space, observation mapping, score function, initial-state distribution, and episode horizon are identical before and after the shift. Only the transition dynamics differ. All offline training data are collected under the dynamics $P_{\mathrm{train}}$, while testing is under dynamics $P_{\mathrm{test}}$.

At time $t$, the environment has a physical state $s_t\in\mathcal{S}$. The agent receives an observation $o_t$, selects an action $a_t\in\mathcal{A}$, and the next state is sampled from $P(\cdot\mid s_t,a_t)$. Each episode lasts at most $T$ steps. The dynamics shift satisfies

\begin{equation}
P_{\mathrm{test}}(\cdot\mid s,a)\neq P_{\mathrm{train}}(\cdot\mid s,a)\quad\text{for some }(s,a)\in\mathcal{S}\times\mathcal{A}. \label{eq:dynamics_shift}
\end{equation}

The task defines a scalar score $r(s)$ that measures task progress. This score is available during offline training and used for evaluation. It is not observed during test-time training or planning. The objective is to recover high performance under $P_{\mathrm{test}}$ by adapting to the changed dynamics. Our experiments instantiate this setting with RGB observations and continuous actions.

\subsection{Action-Conditioned JEPA World Model and Planning}
\label{sec:model_planning}

We use an action-conditioned JEPA world model \citep{lecun2022path, maes2026leworldmodel} with a visual encoder $E_\phi$ and latent dynamics predictor $F_\theta$. The trainable predictor parameters $\theta$ include the action encoder, autoregressive predictor, and prediction projection. Observations are separated by a frame skip of $K$ environment steps, and the $K$ actions between consecutive observations are grouped into one action block. For a sampled trajectory, let $x_k$ denote the observation at frame index $k$, let $b_k$ denote the corresponding action block, and let $z_k=E_\phi(x_k)$ be the encoded latent. Given up to $C$ previous latent states and action blocks, $F_\theta$ predicts the next latent state as $\widehat z_{k+1}=F_\theta\!\left(z_{k-C+1:k},b_{k-C+1:k}\right)$.

Because test-time planning receives neither a goal image nor an online environment reward, we train a reward head before deployment. The reward head maps frozen encoder features to the task score using observation-score pairs collected under the training dynamics. Let $\mathcal{D}_{\mathrm{train}}^r$ denote this offline dataset of pairs $(o,y)$, where $y=r(s)$:
\begin{equation}
    \omega^\star=\arg\min_\omega\;\mathbb{E}_{(o,y)\sim\mathcal{D}_{\mathrm{train}}^r}\!\left[\left(R_\omega(E_\phi(o))-y\right)^2\right].
    \label{eq:reward_head}
\end{equation}
Both the visual encoder $E_\phi$ and the fitted reward head $R_{\omega^\star}$ remain frozen throughout test time. Freezing the encoder gives the test-time training objective a stable representation space, while freezing the reward head keeps the planning objective fixed as the dynamics predictor changes. Keeping the reward head fixed separates adaptation of environment dynamics from adaptation of the task objective. The planner must therefore improve by predicting the consequences of actions more accurately rather than by changing how predicted states are valued.

At each decision step, Cross-Entropy Method (CEM) \citep{rubinstein2004cross} planning samples candidate sequences of $H$ action blocks. The planning horizon spans $HK$ environment steps. The current dynamics predictor rolls each candidate forward autoregressively in latent space, and the frozen reward head scores the predicted trajectory. For a candidate sequence $\mathbf{b}=(b_k,\ldots,b_{k+H-1})$, the planner maximizes the discounted sum of predicted return:
\begin{equation}
    \widehat J_k(\mathbf{b})=\sum_{h=1}^{H}\gamma^{h}R_{\omega^\star}\!\left(\widehat z_{k+h}(\mathbf{b},\theta_k)\right).
    \label{eq:planning_objective}
\end{equation}
CEM repeatedly refits its action distribution toward the highest-scoring candidates. The agent executes only the first action block of the optimized sequence, observes the updated state, and replans. Test-time training therefore changes planning only through the dynamics prediction. The encoder and reward head remain fixed.

\begin{figure}[t]
    \centering
    \includegraphics[width=\textwidth]{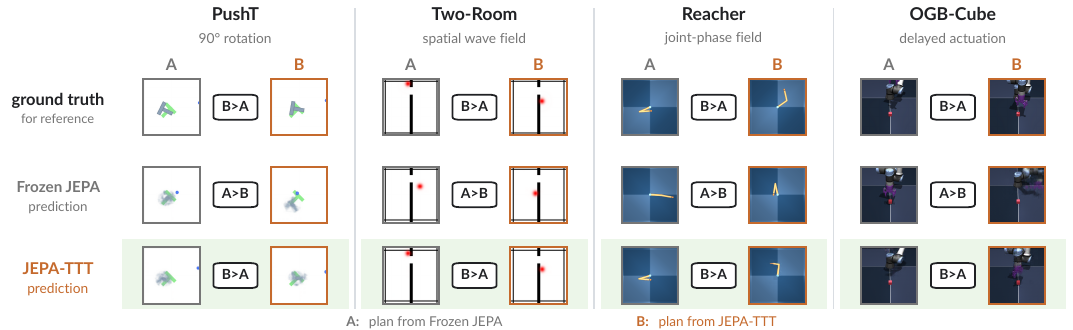}
    \caption{Test-time training can correct the ranking of candidate plans under dynamics shift. For each environment, Plan A is optimized using the frozen JEPA world model, and Plan B is optimized using \method{} from the same initial state. The frozen JEPA world model predicts that A will achieve a higher task score, but the ground-truth rollouts show that B is better. After adaptation, \method{} predicts the correct ordering. Both planners use the same frozen reward head $R_{\omega^\star}$, so the change in plan ranking comes from the adapted latent dynamics predictor. A pretrained decoder is used only to visualize latent predictions.}
    \label{fig:cem_rank_flip}
\end{figure}

Dynamics prediction error leads to incorrect planning because it changes the ordering of candidate actions. Figure~\ref{fig:cem_rank_flip} illustrates this failure mode. Under each shown shift, the frozen JEPA world model assigns the higher predicted return to a plan that performs worse in the environment, while the adapted predictor recovers the correct ordering. These examples are qualitative illustrations. Sections~\ref{sec:planning_results} and \ref{sec:prediction_results} provide the quantitative planning and prediction evaluations.

\subsection{\method{}: Persistent Self-Supervised Test-Time Training}
\label{sec:persistent_ttt}

During test-time interaction, the agent stores its observations and executed actions. Each training window is encoded by the frozen visual encoder to produce latent prediction targets. For window $w$, let $(z_{w,0},\ldots,z_{w,C})$ denote the encoded latents and let $(b_{w,0},\ldots,b_{w,C-1})$ denote the corresponding action blocks. The predictor then rolls forward from the preceding latent context, conditioned on the executed actions. At prediction position $j$, the model estimates $\widehat z_{w,j+1}(\theta)$ from $z_{w,0:j}$ and $b_{w,0:j}$. It is trained to match the frozen encoder representation $z_{w,j+1}$ of the observation. For a minibatch $\mathcal{B}$ and latent dimension $d_z$, we minimize:

\begin{equation}
    \mathcal{L}_{\mathrm{TTT}}(\theta,\mathcal{B})=\frac{1}{|\mathcal{B}|C d_z}\sum_{w\in\mathcal{B}}\sum_{j=0}^{C-1}\left\|\widehat z_{w,j+1}(\theta)-\operatorname{sg}(z_{w,j+1})\right\|_2^2.
    \label{eq:ttt_loss}
\end{equation}

Here, $\operatorname{sg}$ denotes the stop-gradient operator. Because $E_\phi$ is frozen, the target representation remains fixed and test-time training updates only $\theta$. We perform at most one AdamW update at each replanning step when a complete minibatch is available. Freezing the encoder also prevents the target representation from drifting during online predictor optimization. The test-time loss itself is intentionally simple. Our focus is on whether persistent self-supervised prediction updates can adapt to dynamics shift, and on how online data should be organized to make those updates effective.

\method{} retains learned dynamics across test-time episodes. The predictor parameters, optimizer state, and online training buffer all persist across episode boundaries. Let $\theta_{e,t}$, $\nu_{e,t}$, and $\mathcal{Q}_{e,t}$ denote these states at interaction index $t$ of episode $e$, where $t=0$ is the state before the first interaction. If episode $e$ ends with predictor state $\theta_{e,T_e}$, optimizer state $\nu_{e,T_e}$, and buffer $\mathcal{Q}_{e,T_e}$, episode $e+1$ begins from exactly these states: $(\theta_{e+1,0},\nu_{e+1,0},\mathcal{Q}_{e+1,0})=(\theta_{e,T_e},\nu_{e,T_e},\mathcal{Q}_{e,T_e})$. Algorithm~\ref{alg:jepa_ttt} summarizes \method{} and Section~\ref{sec:dense_replay} defines its dense replay update rule.

\begin{algorithm}[t]
\caption{JEPA-TTT}
\label{alg:jepa_ttt}
\KwIn{Pretrained predictor $F_{\theta_0}$, frozen reward head $R_{\omega^\star}$, batch size $B$}
Initialize once: $\theta \leftarrow \theta_0$, replay buffer $\mathcal Q \leftarrow \varnothing$\;

\For{each replanning step in test episodes}{
    $\mathbf b^\star \leftarrow \mathrm{CEM}(F_\theta, R_{\omega^\star})$\;
    Execute the first action block $b_0^\star$\;
    Add all new dense windows to $\mathcal Q$\;
    \If{a scheduled full minibatch is available}{
        Sample $\mathcal B$ uniformly from $\mathcal Q$\;
        Update $\theta$ using $\mathcal L_{\mathrm{TTT}}(\theta,\mathcal B)$\;
    }
}
\end{algorithm}

\subsection{Dense Replay}
\label{sec:dense_replay}

Experience replay reuses online interaction data \citep{lin1992self}. Reactor extends prioritized experience replay to overlapping sequences \citep{gruslys2017reactor}. \method{} uses dense replay to train from accumulated online experience. It forms prediction windows at every valid temporal offset, retains them in a growing buffer, and uniformly samples minibatches for predictor updates.

Dense replay starts a new window at every environment step. Let $a_{u:v}=(a_u,\ldots,a_v)$. For an episode trajectory, the window beginning at environment step $t$ contains $C+1$ observations and the $C$ action blocks between them: $\mathcal{W}_t=\left((o_{t+jK})_{j=0}^{C},\;(a_{t+jK:t+(j+1)K-1})_{j=0}^{C-1}\right).$ Each $\mathcal W_t$ is one training window, and $\mathcal Q$ stores the valid windows accumulated across test-time episodes.

Replay capacity and the ratio of learning updates can both affect performance \citep{fedus2020revisiting}, motivating the following controlled comparisons. Dense construction begins a window at every environment step, whereas sparse construction begins only at action-block boundaries. Thus, dense construction includes all $K$ possible offsets within each action block and yields approximately $K$ times as many windows. The \emph{sparse stream} processes sparse windows chronologically and uses each once. The \emph{compute-matched sparse stream} repeats new sparse windows to match the dense presentation budget without expanding temporal coverage. The \emph{dense stream} processes all dense windows chronologically and uses each once. Dense replay instead samples uniformly from all dense windows accumulated in the buffer.

The comparisons isolate the three factors in sequence: sparse versus compute-matched sparse tests additional optimization; compute-matched sparse versus dense stream tests temporal coverage; and dense stream versus dense replay tests replay from the same available dense window set under a matched cumulative update budget. Appendix~\ref{sec:appendix_test_protocol} specifies the detailed window validity and minibatch scheduling.

%% file: sections/experiments.tex
\begin{figure}[!t]
    \centering
    \includegraphics[width=\textwidth]{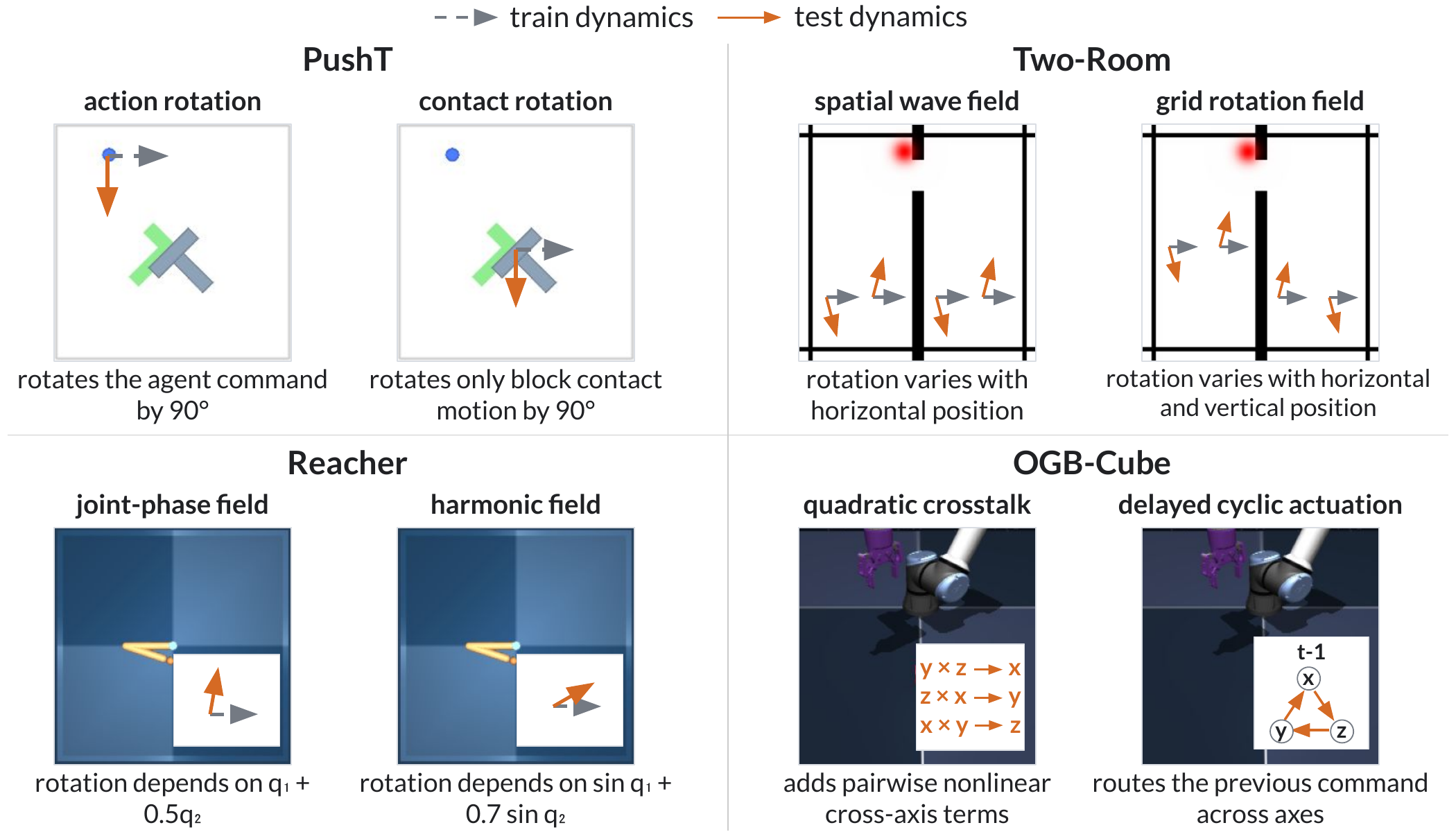}
    \caption{Train and test dynamics illustrated for the eight shifts in our evaluation suite.}
    \label{fig:dynamics_shift_overview}
\end{figure}

\section{Experiments}
\label{sec:experiments}

Our experiments address three research questions:
\begin{itemize}
    \item \textbf{RQ1: Planning recovery.} Does \method{} recover planning performance under dynamics shifts?
    \item \textbf{RQ2: Predictive accuracy.} Does the adapted predictor become more accurate on state prediction?
    \item \textbf{RQ3: Ablation studies.} How do temporal coverage, replay, and persistence affect planning performance?
\end{itemize}

\subsection{Experimental Setup}
\label{sec:experimental_setup}

\paragraph{Tasks and shifts.} We evaluate four diverse continuous-control environments: PushT~\citep{florence2022implicit, chi2025diffusion}, Two-Room~\citep{sobal2025stress}, Reacher from the DeepMind Control Suite~\citep{tassa2018deepmind}, and OGBench-Cube~\citep{park2025ogbench}. All four were also used in LeWorldModel~\citep{maes2026leworldmodel}. Each environment has a fixed task objective with a score in $[0,1]$. For every task, we design two test-time dynamics shifts while keeping the observation mapping, task score, action space, and episode horizon unchanged. The resulting eight shifts cover changes in actuator mapping, contact response, state-dependent control transformations, cross-axis coupling, and delayed actuation.
Figure~\ref{fig:dynamics_shift_overview} illustrates the shifts, and Appendix~\ref{sec:appendix_shifts} gives their detailed definitions.

\paragraph{Offline training.} For each task, we collect 3,000 complete episodes in the training environment and pretrain one action-conditioned JEPA world model (i.e., LeWorldModel). We then freeze its encoder and fit a reward head from offline score labels collected under the same dynamics. Appendix~\ref{sec:appendix_pretraining} provides the data collection, architecture, and optimization details.

\paragraph{Test-time protocol.} Each \method{} deployment consists of 500 episodes under one fixed set of test-time dynamics. At episodes $0, 50, \ldots, 500$, we evaluate the current model on the same fixed set of 100 held-out episodes. We report means and standard deviations over three deployment runs with different environment seeds. Appendix~\ref{sec:appendix_test_protocol} provides the complete test-time training and planning settings.

\paragraph{Metrics.} We report the best held-out planning score (best score), defined as the maximum three-run mean across evaluations at episodes $50, 100, \ldots, 500$. We also report the normalized area under the held-out planning-score curve, averaged over three runs (mean AUC). For each run, we compute the area from episode 0 to 500 by trapezoidal integration and divide by 500. Aggregate results average both metrics across shifts. The best score measures the strongest planning performance reached, while mean AUC summarizes performance throughout test-time training. For dynamics prediction, we report five-block autoregressive latent MSE on fixed trajectories collected under the test-time dynamics. Appendix~\ref{sec:appendix_prediction} details trajectory collection and error aggregation.

\paragraph{Baselines.} Frozen JEPA uses the same pretrained LeWorldModel, reward head, and CEM planner as \method{}, but does not update the world model at test time. PPO-TTT tests whether a pretrained PPO policy can adapt from oracle environment rewards in the same 500-episode interactions \citep{schulman2017proximal}. AdaJEPA is the concurrent work that also adapts a JEPA world model from online transitions within each episode \citep{wang2026adajepa}. For a controlled comparison, we adapt AdaJEPA to our goal-image-free setting, initialize it from the same pretrained LeWorldModel, and use the same frozen reward head and CEM planner as \method{}.

\begin{figure}[t]
    \centering
    \includegraphics[width=\textwidth]{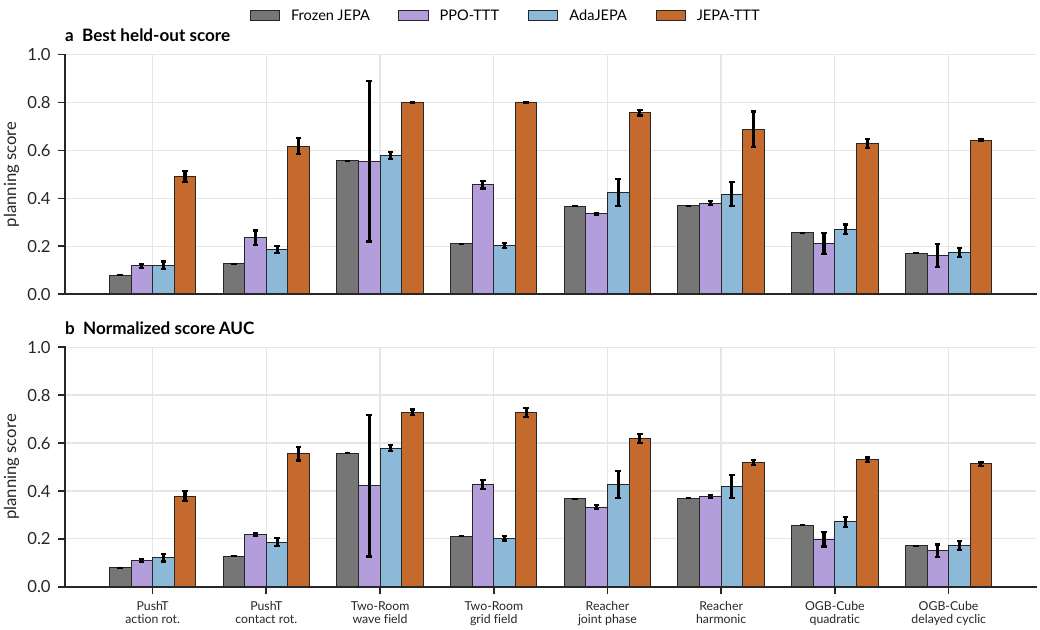}
    \caption{Planning performance under the test-time dynamics shift. Panel (a) reports the best held-out score. Panel (b) reports normalized held-out planning-score AUC from episode 0 to 500. Bars and error bars report means and standard deviations over three deployment runs. \method{} outperforms other methods including the frozen JEPA world model baseline, PPO test-time training (PPO-TTT), and AdaJEPA.}
    \label{fig:main_baseline_comparison}
\end{figure}

\begin{figure}[p]
    \centering
    \includegraphics[width=0.977\textwidth]{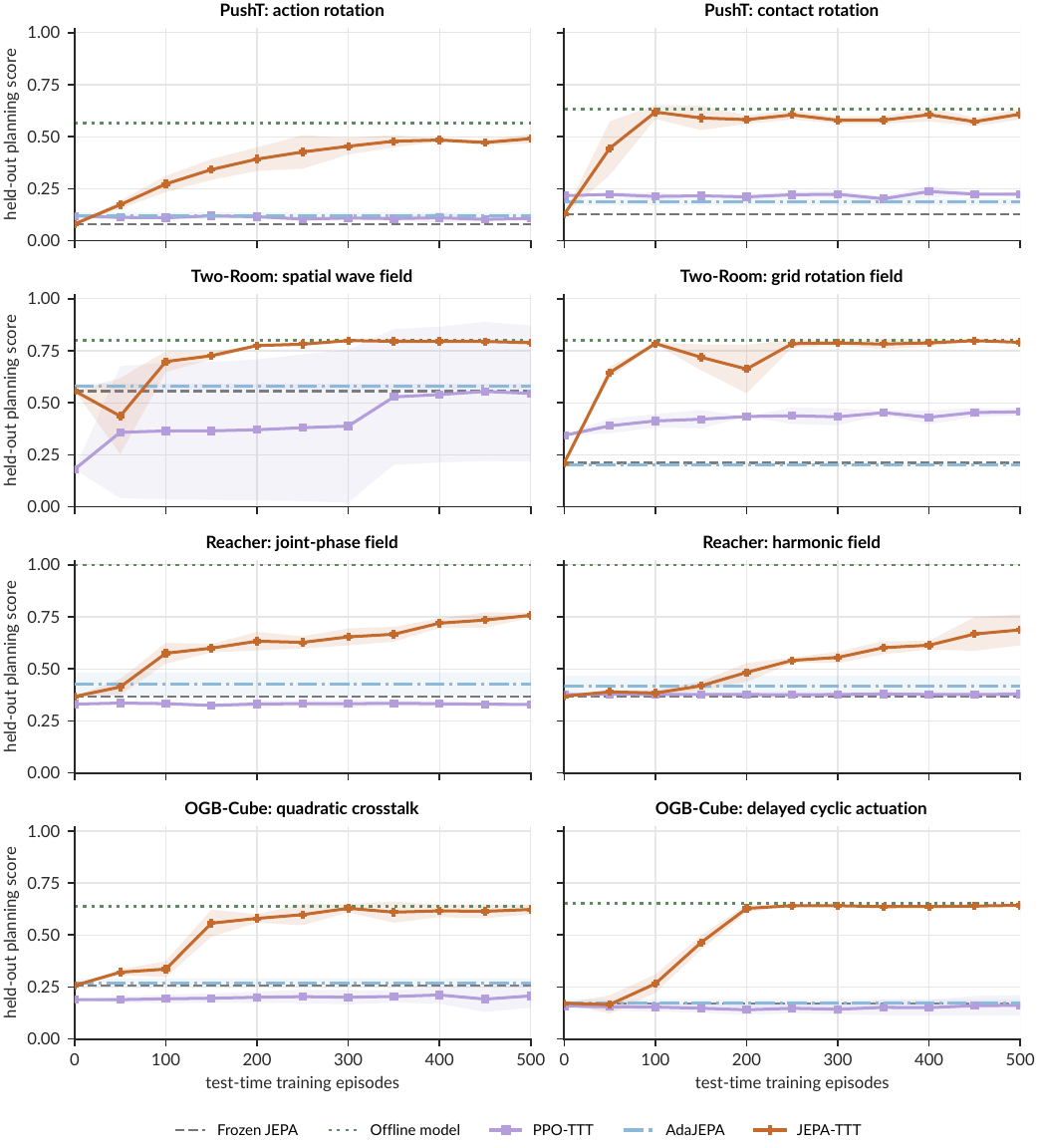}
    \caption{Held-out planning performance throughout 500 test-time training episodes. Curves show means over three deployment runs evaluated on the same fixed set of 100 held-out episodes, and shaded regions show standard deviations. \method{} is very close to, or even slightly surpasses, the offline reference using only 500 episodes of online interaction, and outperforms other methods including the frozen JEPA world model baseline, PPO test-time training (PPO-TTT), and AdaJEPA.}
    \label{fig:main_capability}
\end{figure}

\subsection[Test-Time Training Improves Planning under Dynamics Shifts]{Test-Time Training Improves Planning under Dynamics Shifts}
\label{sec:planning_results}

For RQ1, \method{} improves planning across all eight dynamics shifts. Relative to the frozen JEPA world model, \method{} improves best score by 153\% and mean AUC by 113\%, with both metrics averaged across shifts. \method{} outperforms all three baselines on every evaluated shift for both reported planning metrics. Figure~\ref{fig:main_baseline_comparison} summarizes these results, and Appendix Table~\ref{tab:baseline_best_auc} reports the exact values.

The large PPO-TTT variance on the Two-Room spatial-wave shift reflects qualitatively different learning outcomes across deployment runs. One run remains trapped in the first room, one discovers the doorway after roughly 350 test episodes, and one discovers it within the first 50 episodes.

The full learning curves in Figure~\ref{fig:main_capability} show that \method{} improves as interaction accumulates. Its best score and mean AUC improve over the frozen JEPA world model on all eight shifts. We include a data-rich offline reference trained directly under the test-time dynamics to contextualize the remaining adaptation gap. It is not a deployable baseline and is not a mathematical upper bound because it uses a different offline data collection and uses 3,000 complete episodes from the shifted environment for 20 epochs. On six of the eight shifts, the best score of \method{} is very close to, or even slightly surpasses, the offline reference using only 500 episodes of online interaction. Appendix Table~\ref{tab:target_domain_reference} reports the reference values.

\subsection{Test-Time Training Improves Dynamics Prediction}
\label{sec:prediction_results}

\begin{figure}[ht]
    \centering
    \includegraphics[width=0.98\textwidth]{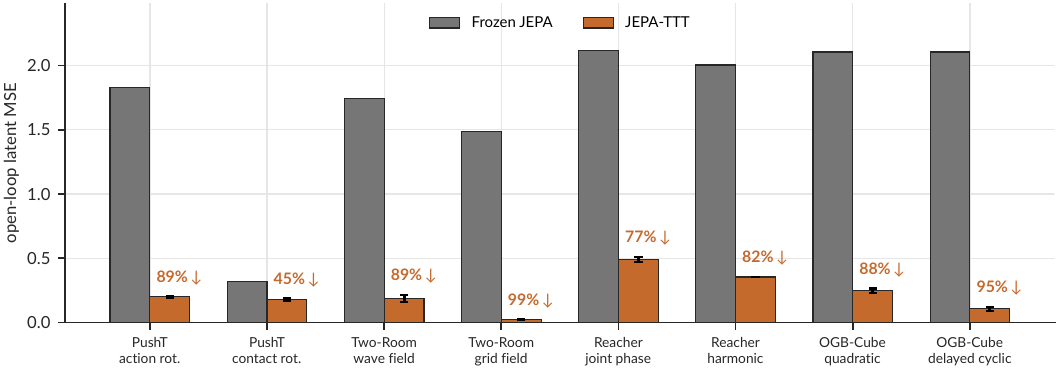}
    \caption{Five-block autoregressive latent prediction error under the test-time dynamics. The orange bars show JEPA-TTT's prediction errors drop significantly after 500 test-time episodes. Percent labels show the relative MSE reduction after adaptation, and error bars show standard deviations over the three adapted models.}
    \label{fig:prediction_recovery}
\end{figure}

For RQ2, we evaluate the frozen JEPA predictor and the adapted predictor on identical held-out trajectories. Both models receive the same observed context and future action sequence, then predict five latent states autoregressively without intermediate observations. \method{} lowers latent prediction MSE on every shift. The mean reduction is 83\% after 500 test-time episodes. These reductions indicate that the predictor has become more accurate under the changed dynamics. Figure~\ref{fig:prediction_recovery} reports the comparison.

The planning and prediction results are therefore consistent with the intended mechanism: \method{} improves dynamics prediction under the shifted dynamics and supports better planning.

\input{sections/dense_ablation_table}

\subsection[Ablation Studies]{Ablation Studies}
\label{sec:ablation_studies}
\label{sec:dense_ablation}

For RQ3, we ablate dense replay by separately varying update count, temporal coverage, and replay. Averaged across shifts, the compute-matched sparse stream improves both best score and mean AUC over the sparse stream, indicating a benefit from additional optimization. The dense stream further improves both metrics under the same presentation budget, suggesting a benefit from broader temporal coverage. Dense replay achieves the highest aggregate best score and mean AUC among the four update rules. Compared with the dense stream, it reaches a similar best score (0.678 versus 0.671) while improving mean AUC (0.571 versus 0.548). Table~\ref{tab:auc_all_methods} reports the full comparison. These results support dense temporal coverage and suggest a small additional benefit from replay.

We also test the effect of retaining adaptation across episodes. In a matched control with episode-local dense windows, retaining the predictor and optimizer state increases the best score from 0.289 to 0.729 and mean AUC from 0.273 to 0.637, with the same number of predictor updates. Appendix Table~\ref{tab:persistence_control} reports the full results.

\FloatBarrier

%% file: sections/dense_ablation_table.tex
\begin{table}[!t]
    \centering
    {\small
    \begin{tabular}{llcccc}
        \toprule
        Shift & Metric & {\makecell{Sparse\\stream}} & {\makecell{Compute-matched\\sparse stream}} & {\makecell{Dense\\stream}} & {\makecell{Dense\\replay}} \\
        \midrule
        \multirow{2}{*}{PushT: action rotation} & Best & $0.253\pm0.063$ & $0.236\pm0.049$ & $\mathbf{0.511\pm0.009}$ & $0.490\pm0.022$ \\
        & AUC & $0.169\pm0.007$ & $0.169\pm0.021$ & $0.359\pm0.044$ & $\mathbf{0.378\pm0.022}$ \\
        \cmidrule(lr){1-6}
        \multirow{2}{*}{PushT: contact rotation} & Best & $0.317\pm0.258$ & $0.521\pm0.319$ & $\mathbf{0.646\pm0.044}$ & $0.619\pm0.033$ \\
        & AUC & $0.205\pm0.082$ & $0.320\pm0.157$ & $0.490\pm0.041$ & $\mathbf{0.555\pm0.028}$ \\
        \cmidrule(lr){1-6}
        \multirow{2}{*}{Two-Room: spatial wave} & Best & $0.505\pm0.031$ & $0.790\pm0.002$ & $0.797\pm0.002$ & $\mathbf{0.799\pm0.002}$ \\
        & AUC & $0.427\pm0.034$ & $0.672\pm0.020$ & $0.697\pm0.012$ & $\mathbf{0.728\pm0.012}$ \\
        \cmidrule(lr){1-6}
        \multirow{2}{*}{Two-Room: grid rotation} & Best & $0.786\pm0.004$ & $0.794\pm0.003$ & $0.782\pm0.000$ & $\mathbf{0.799\pm0.003}$ \\
        & AUC & $0.662\pm0.009$ & $0.717\pm0.026$ & $0.686\pm0.020$ & $\mathbf{0.725\pm0.019}$ \\
        \cmidrule(lr){1-6}
        \multirow{2}{*}{Reacher: joint phase} & Best & $0.557\pm0.025$ & $\mathbf{0.768\pm0.062}$ & $0.725\pm0.018$ & $0.758\pm0.012$ \\
        & AUC & $0.401\pm0.007$ & $0.545\pm0.051$ & $0.593\pm0.013$ & $\mathbf{0.619\pm0.019}$ \\
        \cmidrule(lr){1-6}
        \multirow{2}{*}{Reacher: harmonic} & Best & $0.446\pm0.046$ & $0.588\pm0.048$ & $0.610\pm0.039$ & $\mathbf{0.688\pm0.075}$ \\
        & AUC & $0.398\pm0.007$ & $0.488\pm0.014$ & $0.492\pm0.006$ & $\mathbf{0.518\pm0.012}$ \\
        \cmidrule(lr){1-6}
        \multirow{2}{*}{OGB-Cube: quadratic} & Best & $0.520\pm0.053$ & $0.652\pm0.010$ & $\mathbf{0.653\pm0.004}$ & $0.628\pm0.020$ \\
        & AUC & $0.390\pm0.005$ & $0.528\pm0.015$ & $0.528\pm0.013$ & $\mathbf{0.530\pm0.009}$ \\
        \cmidrule(lr){1-6}
        \multirow{2}{*}{OGB-Cube: delayed cyclic} & Best & $0.591\pm0.044$ & $0.646\pm0.006$ & $\mathbf{0.646\pm0.004}$ & $0.643\pm0.003$ \\
        & AUC & $0.290\pm0.022$ & $0.522\pm0.006$ & $\mathbf{0.542\pm0.006}$ & $0.513\pm0.009$ \\
        \midrule
        \multirow{2}{*}{Mean} & Best & $0.497\pm0.028$ & $0.625\pm0.025$ & $0.671\pm0.008$ & $\mathbf{0.678\pm0.013}$ \\
        & AUC & $0.368\pm0.007$ & $0.495\pm0.025$ & $0.548\pm0.007$ & $\mathbf{0.571\pm0.004}$ \\
        \bottomrule
    \end{tabular}
    }
    \caption{Comparisons of different update rules for test-time training. The sparse stream uses its natural number of updates. Entries report means and standard deviations over three runs. The Mean rows summarize the three run-level averages across shifts. Bold marks the highest unrounded mean for each metric and shift. Dense replay has the highest aggregate mean best score and AUC.}
    \label{tab:auc_all_methods}
\end{table}

%% file: sections/conclusion.tex
\section{Conclusion}
\label{sec:conclusion}

We studied whether a pretrained action-conditioned JEPA world model can adapt its dynamics model during deployment. \method{} updates only the latent dynamics predictor from observed transitions while keeping the visual encoder and reward head fixed. The adapted predictor is retained across episodes, allowing adaptation to accumulate over time. Across eight dynamics shifts in four continuous-control environments, \method{} improves planning on every shift. The planning gains are accompanied by lower latent prediction error. Ablation studies show the benefit of dense temporal windows, with a small further gain from replay.

\section{Limitations}
\label{sec:limitations}

\paragraph{Scope of dynamics shifts.} We study a single dynamics change that remains fixed throughout test time. The visual observation mapping, task score, action interface, and episode horizon do not change. Because the encoder and reward head remain frozen, \method{} is not designed to handle simultaneous visual or reward shifts.

\paragraph{Continual adaptation and catastrophic forgetting.} We do not study sequences of dynamics changes or a return to the training dynamics after adaptation. Persistent predictor updates may degrade predictions for previously learned dynamics, including the training dynamics. Our experiments therefore do not measure retention, forward/backward transfer, or catastrophic forgetting in continual learning settings.

%% file: sections/appendix.tex
\clearpage
\beginsupplement

\begin{center}
    {\sffamily\fontseries{eb}\Large\selectfont Appendix}
\end{center}

\section{Experimental Protocol and Implementation}
\label{sec:appendix_protocol}

\subsection{World-Model Pretraining}
\label{sec:appendix_pretraining}

All observations are $224 \times 224$ RGB images. All four world models use the same LeWorldModel configuration. The image encoder is a ViT-Tiny with $14\times14$ patches and 192-dimensional outputs. The autoregressive predictor has six transformer layers, 16 attention heads of dimension 64, hidden dimension 192, MLP dimension 2048, and dropout 0.1. The action encoder maps one five-action block to 192 dimensions. The latent and prediction projections are two-layer MLPs with hidden dimension 2048 and BatchNorm. Each sample contains three context keyframes and one prediction keyframe separated by five raw simulator steps.

We collect 3,000 complete episodes under the training dynamics for each task. PushT uses an equal mixture of block-local and arena-wide target actions held for four steps. Two-Room uses the upstream noisy heuristic, Reacher uses random actions, and OGB-Cube uses an equal mixture of guided and Gaussian exploration, with 30\% uniform actions during guided episodes. Maximum collection horizons are 300, 100, 200, and 50 steps, respectively. These collection distributions do not use task rewards and need not equal the fixed-task evaluation reset distribution. Models are initialized from scratch and trained for 20 epochs using AdamW, batch size 128, maximum learning rate $5\times10^{-5}$, weight decay $10^{-3}$, gradient clipping 1.0, linear warmup followed by cosine decay, and bfloat16 precision. We use a random 90/10 window split and the predeclared epoch-20 checkpoint rather than selecting an epoch on test performance. The upstream LeWorldModel latent prediction and SIGReg objectives are unchanged. SIGReg uses weight 0.09, 17 knots, and 1,024 projections.

After pretraining, the reward head is fitted on an episode-disjoint split collected under the training dynamics. It is a sigmoid MLP with dimensions $192\rightarrow256\rightarrow256\rightarrow1$ and ReLU hidden activations. Training uses AdamW, batch size 512, learning rate $10^{-3}$, weight decay $10^{-5}$, and early stopping with patience 15 for at most 100 epochs. The selected epochs are 67, 64, 82, and 61 for PushT, Two-Room, Reacher, and OGBench-Cube. The optional decoder is trained for 20 epochs with AdamW, batch size 32, learning rate $2\times10^{-4}$, weight decay $10^{-4}$, gradient clipping 1.0, and bfloat16 autocast. Neither the decoder nor its output is used for planning.

\subsection{Planning and Test-Time Training}
\label{sec:appendix_test_protocol}

\begin{table}[!htbp]
    \centering
    {\small
    \renewcommand{\arraystretch}{1.12}
    \begin{tabular}{ll}
        \toprule
        Component & Setting \\
        \midrule
        Test-time interaction horizon & 500 episodes under the test-time dynamics \\
        Episode horizon & PushT: 300 raw steps, other tasks: 50 \\
        Held-out evaluations & episodes $0,50,\ldots,500$ \\
        Evaluation size & 100 held-out episodes per evaluation \\
        CEM & 256 candidates, 6 iterations, 32 elites \\
        Planning horizon & 5 world-model blocks \\
        Execution & 1 block, then replan \\
        Action block & 5 raw actions \\
        CEM proposal & zero-centered Gaussian, initial std. 1.0, std. floor 0.05 \\
        Reward discount & 0.99 \\
        Reward-head scale & PushT: 10, other tasks: 1 (rank-invariant for CEM) \\
        Test-time optimizer & AdamW, LR $5\times10^{-5}$, WD $10^{-3}$ \\
        Test-time batch size & 64 \\
        Gradient clipping & 1.0 \\
        Trainable-module mode & evaluation mode; normalization statistics remain fixed \\
        Update rate & at most 1 update per replan \\
        Trainable components & action encoder, predictor, prediction projection \\
        Frozen components & image encoder, latent projector, reward head, decoder \\
        Test-time inputs & current RGB observations and executed actions \\
        \bottomrule
    \end{tabular}
    }
    \caption{Shared test-time planning and training settings. These settings define the common test-time protocol for all eight shifts.}
    \label{tab:protocol_hyperparameters}
\end{table}

For dense construction, a window start becomes valid only after a clip spanning $(C+1)K$ raw observation rows from that start is available. The loss uses the keyframes and first $C$ action blocks in $\mathcal{W}_t$. Sparse construction creates its first window after $C$ complete action blocks. For the compute-matched sparse stream, each new sparse window is repeated $N_\tau-N_{\tau-1}$ times, where $N_\tau$ is the number of dense windows available after replanning step $\tau$.

Test-time interaction continues to the fixed episode horizon after success so that data supply does not depend on the current model. Held-out evaluation episodes use normal task termination. The sparse stream, compute-matched sparse stream, and dense stream update only when a full batch is available. Windows that do not complete a batch remain pending at scheduled evaluations and at episode 500 rather than being flushed as a smaller batch. Dense replay receives the same number of updates as the dense stream. After $N$ unique dense windows, both have $\lfloor N/64\rfloor$ updates. Each replay batch samples without replacement within that batch, while all stored windows remain valid for later updates.

For \method{} and each dense replay ablation, all three deployment runs begin from the same fixed pretrained world model. The runs differ only in their test-time environment-seed sequences and therefore in the interaction data accumulated during adaptation. Held-out evaluation uses a separate fixed seed suite that is disjoint from all deployment runs, and evaluation trajectories never enter the online buffer. The reported standard deviations therefore characterize variability in test-time interaction and adaptation conditional on one pretrained model. They do not measure variability across independently pretrained world models.

\section{Tasks and Dynamics Shifts}
\label{sec:appendix_shifts}

Let
\begin{equation}
    \mathbf{R}(\alpha)=
    \begin{bmatrix}
        \cos\alpha & -\sin\alpha \\
        \sin\alpha & \cos\alpha
    \end{bmatrix}
\end{equation}
denote a planar rotation. Every shift below changes only how the commanded action produces the next state. The visual goal, score, action bounds, and initial-state distribution are identical to the corresponding task under the training dynamics.

\subsection{PushT}

The action is a normalized two-dimensional target for the circular agent. The instantaneous score is the overlap coverage between the movable T block and the fixed green target, and the reported episode score is its peak value. Under \emph{action rotation}, a target $u_t$ is rotated by $90^\circ$ around the current agent position $p_t$ before the base controller receives it:
\begin{equation}
    u_t^{\mathrm{apply}}=p_t+\mathbf{R}(\pi/2)(u_t-p_t).
\end{equation}
Under \emph{contact rotation}, the agent follows the commanded target normally. At each physics substep, only the contact-induced block displacement $\Delta q$ is replaced by $\mathbf{R}(\pi/2)\Delta q$. This separates an actuator-frame shift from a contact-response shift.

\subsection{Two-Room}

The agent begins in the left room and moves toward a fixed goal $g=(164,112)$ in pixel coordinates. Let $d=(112,49)$ be the doorway waypoint. The score is $\exp(-D(s,g)/64)$, where $D(s,g)=\lVert s-d\rVert_2+\lVert d-g\rVert_2$ while the agent is left of the wall and $D(s,g)=\lVert s-g\rVert_2$ after it crosses. A run succeeds within 16 pixels of $g$.

Both shifts apply $a_t^{\mathrm{apply}}=\mathbf{R}(\alpha(s_t))a_t$. For the spatial wave field,
\begin{equation}
    \alpha(s)=75^\circ\sin\!\left(\frac{2\pi(x-20)}{88}\right).
\end{equation}
For the grid field,
\begin{equation}
    \alpha(s)=75^\circ\sin\!\left(\frac{2\pi(x-20)}{96}\right)\cos\!\left(\frac{2\pi(y-20)}{96}\right).
\end{equation}
The first field changes smoothly with horizontal position; the second changes over both visible room coordinates.

\subsection{Reacher}

The goal is the fixed joint configuration $(0,0)$. Let $\widetilde q\in[-\pi,\pi)^2$ be the wrapped joint-angle error. The score is $\exp(-\operatorname{mean}(\widetilde q^2)/2)$, and success requires $|\widetilde q_i|<0.05$ for both joints. The shifts rotate the two-dimensional motor command by an angle that depends on the current configuration and vanishes near the goal.

Define
\begin{equation}
    h(\widetilde q)=1-\exp\!\left(-\frac{\operatorname{mean}(\widetilde q^2)}{2(0.20)^2}\right), \qquad \alpha(\widetilde q)=\alpha_{\max}h(\widetilde q)\tanh\!\left(\frac{\psi(\widetilde q)}{\kappa}\right).
\end{equation}
The joint-phase field uses $\psi(\widetilde q)=\widetilde q_1+0.5\widetilde q_2$, $\kappa=0.35$, and $\alpha_{\max}=155^\circ$. The harmonic field uses $\psi(\widetilde q)=\sin\widetilde q_1+0.7\sin\widetilde q_2$, $\kappa=0.30$, and $\alpha_{\max}=165^\circ$. In both cases, $a_t^{\mathrm{apply}}=\operatorname{clip}(\mathbf{R}(\alpha(\widetilde q_t))a_t,-1,1)$.

\subsection{OGBench-Cube}

We retain the OGBench-Cube simulator, UR5e robot, camera, table, and physical red cube, and use a three-dimensional reach task. The target is a fixed hover point 0.10 m above the cube at $(0.425,0,0.12)$ m. The action contains Cartesian $x,y,z$ commands; yaw and gripper commands are fixed at zero. For end-effector error $e$, the score is $\exp(-\lVert e\rVert_2/0.10)$ and success requires $\lVert e\rVert_2\leq0.05$ m.

For quadratic crosstalk, a command $a=(a_x,a_y,a_z)$ becomes
\begin{equation}
    a^{\mathrm{apply}}=\operatorname{clip}\!\left(a+3(a_ya_z,a_za_x,a_xa_y),-1,1\right).
\end{equation}
For delayed cyclic actuation, the command applied at raw step $t$ is
\begin{equation}
    a_t^{\mathrm{apply}}=(a_{t-1,y},a_{t-1,z},a_{t-1,x}),
\end{equation}
with the previous command initialized to zero at reset. This combines a one-step delay with a fixed cross-axis routing.

\section{Baseline Details}
\label{sec:appendix_baselines}

\paragraph{AdaJEPA.} We evaluate AdaJEPA's episodic adaptation procedure using the same pretrained model, reward head, and CEM planner in \method{} without a goal image. We do not use its original goal-conditioned planning formulation to ensure fairness. Each episode restores the pretrained parameters and clears the history of the five most recent executed action segments. Every replanning step initializes a new Adam optimizer and performs one update. All adaptation is discarded at the end of the episode. The predictor learning rate is $5\times10^{-4}$ and the encoder-head rate, when adapted, is $10^{-5}$, without weight decay or gradient clipping. We evaluate adaptation on the last layer of the predictor, either with a frozen encoder head (pred-last+enc-frozen) or with the default trainable projection head (pred-last+enc-last). We evaluate AdaJEPA in three runs of 100 episodes. Before every episode, AdaJEPA reloads the pretrained model, adapts only from transitions observed during that episode, and produces one task score. We average the 100 episode scores within each run, then report the mean and sample standard deviation across the three run averages. Because no adaptation carries between episodes, we use this per-episode performance estimate for both best score and mean AUC.

\paragraph{PPO-TTT.} Base PPO policies use a CNN encoder with 32, 64, and 64 channels, followed by a 256-dimensional feature layer and diagonal-Gaussian actor and value heads. PushT requested 10 million training steps, stopped after 2.40 million by a predeclared validation-plateau rule, and selected the checkpoint at 1.30 million steps. Two-Room, Reacher, and OGB-Cube each trained for 2 million steps and selected checkpoints at 1.5, 1.2, and 1.7 million steps, respectively, using fixed held-out validation episodes. The selected policy, value function, and a fresh optimizer then persist across 500 test episodes and receive the online environment reward. Every 10 episodes, Adam runs 10 epochs with learning rate $3\times10^{-4}$, minibatch size 512, discount 0.99, GAE parameter 0.95, clipping coefficient 0.2, value coefficient 0.5, entropy coefficient 0, and gradient clipping 0.5. Test-time actions are stochastic. Held-out evaluation actions are deterministic.

\section{Complete Planning Results}
\label{sec:appendix_results}

\begin{table}[H]
    \centering
    {\small
    \begin{tabular}{llcccc}
        \toprule
        Shift & Metric & Frozen JEPA & PPO-TTT & AdaJEPA & \method{} \\
        \midrule
        \multirow{2}{*}{PushT: action rotation} & Best & 0.080 & $0.119\pm0.009$ & $0.121\pm0.015$ & $\mathbf{0.490\pm0.022}$ \\
        & AUC & 0.080 & $0.110\pm0.006$ & $0.121\pm0.015$ & $\mathbf{0.378\pm0.022}$ \\
        \cmidrule(lr){1-6}
        \multirow{2}{*}{PushT: contact rotation} & Best & 0.127 & $0.237\pm0.029$ & $0.186\pm0.016$ & $\mathbf{0.619\pm0.033}$ \\
        & AUC & 0.127 & $0.219\pm0.005$ & $0.186\pm0.016$ & $\mathbf{0.555\pm0.028}$ \\
        \cmidrule(lr){1-6}
        \multirow{2}{*}{Two-Room: spatial wave} & Best & 0.557 & $0.554\pm0.334$ & $0.579\pm0.013$ & $\mathbf{0.799\pm0.002}$ \\
        & AUC & 0.557 & $0.421\pm0.295$ & $0.579\pm0.013$ & $\mathbf{0.728\pm0.012}$ \\
        \cmidrule(lr){1-6}
        \multirow{2}{*}{Two-Room: grid rotation} & Best & 0.211 & $0.456\pm0.015$ & $0.203\pm0.010$ & $\mathbf{0.799\pm0.003}$ \\
        & AUC & 0.211 & $0.426\pm0.019$ & $0.203\pm0.010$ & $\mathbf{0.725\pm0.019}$ \\
        \cmidrule(lr){1-6}
        \multirow{2}{*}{Reacher: joint phase} & Best & 0.367 & $0.336\pm0.005$ & $0.426\pm0.056$ & $\mathbf{0.758\pm0.012}$ \\
        & AUC & 0.367 & $0.332\pm0.009$ & $0.426\pm0.056$ & $\mathbf{0.619\pm0.019}$ \\
        \cmidrule(lr){1-6}
        \multirow{2}{*}{Reacher: harmonic} & Best & 0.369 & $0.379\pm0.009$ & $0.418\pm0.049$ & $\mathbf{0.688\pm0.075}$ \\
        & AUC & 0.369 & $0.377\pm0.007$ & $0.418\pm0.049$ & $\mathbf{0.518\pm0.012}$ \\
        \cmidrule(lr){1-6}
        \multirow{2}{*}{OGB-Cube: quadratic} & Best & 0.257 & $0.212\pm0.043$ & $0.271\pm0.020$ & $\mathbf{0.628\pm0.020}$ \\
        & AUC & 0.257 & $0.199\pm0.031$ & $0.271\pm0.020$ & $\mathbf{0.530\pm0.009}$ \\
        \cmidrule(lr){1-6}
        \multirow{2}{*}{OGB-Cube: delayed cyclic} & Best & 0.172 & $0.162\pm0.049$ & $0.173\pm0.019$ & $\mathbf{0.643\pm0.003}$ \\
        & AUC & 0.172 & $0.151\pm0.025$ & $0.173\pm0.019$ & $\mathbf{0.513\pm0.009}$ \\
        \midrule
        \multirow{2}{*}{Mean} & Best & 0.267 & 0.307 & 0.297 & \textbf{0.678} \\
        & AUC & 0.267 & 0.279 & 0.297 & \textbf{0.571} \\
        \bottomrule
    \end{tabular}
    }
    \caption{Exact values for Figure~\ref{fig:main_baseline_comparison}. \method{} and PPO-TTT's entries are means and sample standard deviations over three deployment runs. For AdaJEPA, each run averages the task scores from 100 episodes, with the pretrained model reloaded before every episode. Entries report the mean and sample standard deviation across three runs. Because AdaJEPA does not retain adaptation across episodes, the same score is reported for both metrics. \method{} outperforms all three baselines on every shift under both metrics.}
    \label{tab:baseline_best_auc}
\end{table}

\begin{figure}[H]
    \centering
    \includegraphics[width=\textwidth]{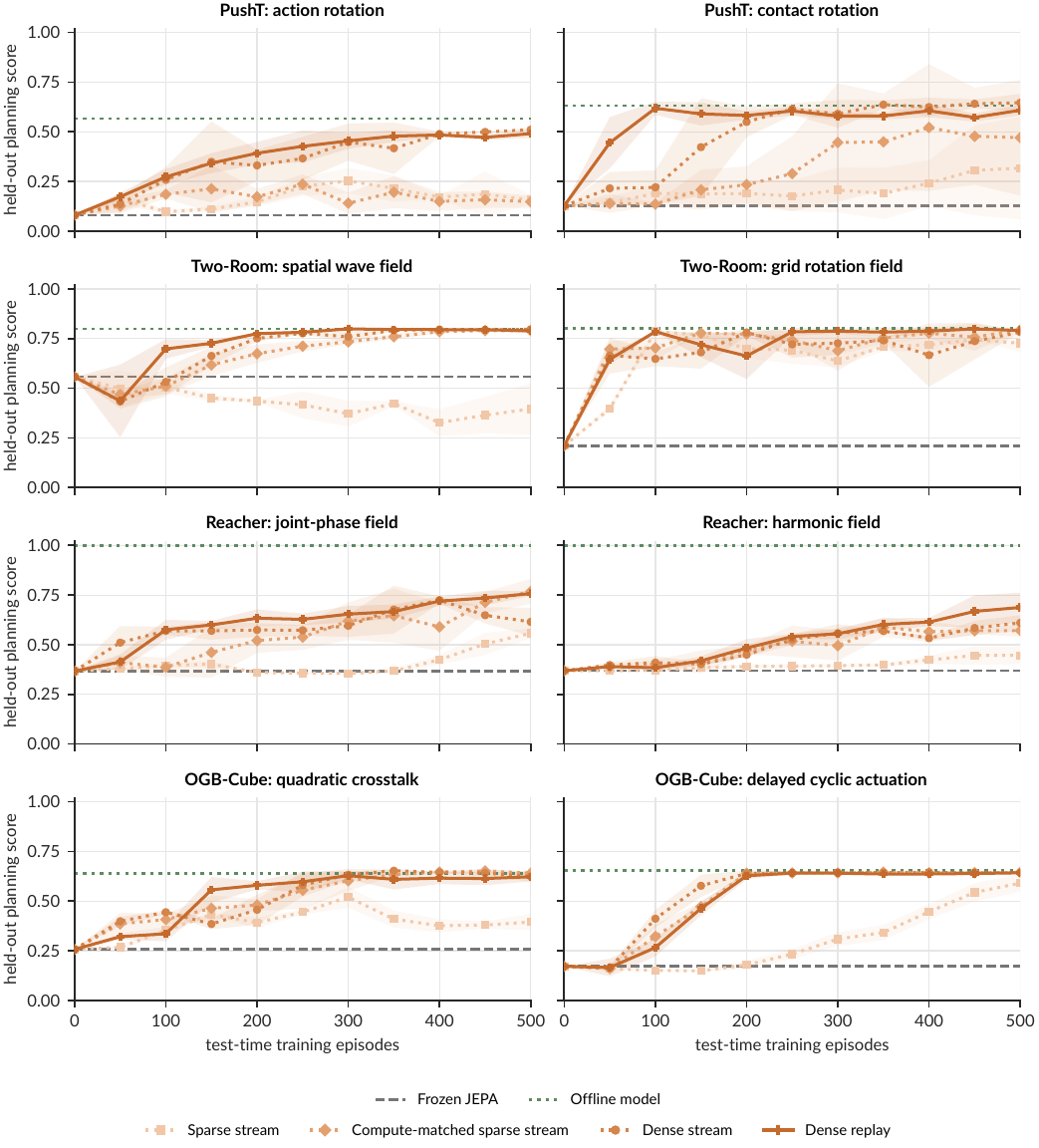}
    \caption{Full held-out score curves for \method{} and the three ablations of dense replay. Curves show means and sample standard deviations over three runs. Dashed gray lines show the frozen JEPA world model. Dotted green lines show the data-rich offline reference trained under the test-time dynamics. This reference is neither a deployable baseline nor a mathematical upper bound because it uses a different data-collection distribution and receives 3,000 complete offline episodes from the shifted environment (Table~\ref{tab:target_domain_reference}). Dense replay achieves the highest aggregate best score and AUC.}
    \label{fig:all_capability}
\end{figure}

\subsection{Trainable Components in AdaJEPA}

\begin{table}[H]
    \centering
    {\small
    \begin{tabular}{lcc}
        \toprule
        Shift & \makecell{pred-last +\\enc-last (default)} & \makecell{pred-last +\\enc-frozen} \\
        \midrule
        PushT: action rotation & $\mathbf{0.1210\pm0.0155}$ & $0.1042\pm0.0163$ \\
        PushT: contact rotation & $\mathbf{0.1864\pm0.0156}$ & $0.1527\pm0.0145$ \\
        \cmidrule(lr){1-3}
        Two-Room: spatial wave & $0.5791\pm0.0134$ & $\mathbf{0.5850\pm0.0125}$ \\
        Two-Room: grid rotation & $0.2031\pm0.0101$ & $\mathbf{0.2470\pm0.0028}$ \\
        \cmidrule(lr){1-3}
        Reacher: joint phase & $0.4259\pm0.0565$ & $\mathbf{0.4266\pm0.0552}$ \\
        Reacher: harmonic & $0.4176\pm0.0490$ & $\mathbf{0.4183\pm0.0487}$ \\
        \cmidrule(lr){1-3}
        OGB-Cube: quadratic & $0.2711\pm0.0199$ & $\mathbf{0.2965\pm0.0144}$ \\
        OGB-Cube: delayed cyclic & $\mathbf{0.1730\pm0.0191}$ & $0.1608\pm0.0204$ \\
        \midrule
        Mean & 0.2972 & \textbf{0.2989} \\
        \bottomrule
    \end{tabular}
    }
    \caption{AdaJEPA performance with two choices of trainable components. Each run averages the task scores from 100 episodes, with the pretrained model reloaded before every episode. Entries report the mean and sample standard deviation across three runs. Because adaptation does not carry across episodes, the same performance estimate is used for both best held-out score and normalized held-out planning-score AUC. Bold marks the higher mean within each shift. The difference between the two configurations is only 0.0017. All main comparisons use the default pred-last+enc-last configuration.}
    \label{tab:adajepa_scope}
\end{table}

\FloatBarrier

\section{Prediction and Decoder Diagnostics}
\label{sec:appendix_prediction}

\subsection{Fixed-Trajectory Latent Error}

Within each row of Table~\ref{tab:prediction_raw}, trajectories are collected once under the test-time dynamics and then reused unchanged for the frozen JEPA world model and the adapted model. They are neither expert demonstrations nor trajectories generated by an adapted model. PushT uses 100 fixed, seed-deterministic 300-step behavior trajectories. Every four simulator steps, this behavior policy samples a target with equal probability either uniformly over the arena or near the movable block with 60-pixel Gaussian jitter. Two-Room, Reacher, and OGB-Cube each use 100 fixed 50-step trajectories generated by CEM using the frozen JEPA world model and reward head. Observation and action hashes verify that every model receives exactly the same inputs.

We extract every sliding window containing three observed context frames and five future frames, with adjacent frames separated by five environment steps. The frozen image encoder maps all RGB observations to target latents. Starting from the three observed context latents, the predictor autoregressively predicts five future latents under the executed actions without receiving intermediate observations. At each horizon, MSE is averaged over the latent dimensions. Windows are averaged within each episode before averaging episodes, so long trajectories do not receive greater weight. Table~\ref{tab:prediction_raw} averages over the five horizons. Adapted values are means and sample standard deviations over the three dense replay models after episode 500. Raw magnitudes should not be compared across tasks because their latent spaces are trained independently.

\begin{table}[H]
    \centering
    {\small
    \begin{tabular}{lcc@{\hspace{1.2em}}c}
        \toprule
        Shift & Frozen JEPA MSE & Adapted MSE & Reduction \\
        \midrule
        PushT action rotation & 1.82540 & $0.19988\pm0.00491$ & 89.05\% \\
        PushT contact rotation & 0.32117 & $0.17751\pm0.00953$ & 44.73\% \\
        Two-Room spatial wave & 1.74541 & $0.18697\pm0.02870$ & 89.29\% \\
        Two-Room grid field & 1.48501 & $0.02205\pm0.00345$ & 98.52\% \\
        Reacher joint phase & 2.11969 & $0.48927\pm0.02162$ & 76.92\% \\
        Reacher harmonic & 2.00316 & $0.35488\pm0.00216$ & 82.28\% \\
        OGB-Cube quadratic & 2.10429 & $0.24754\pm0.01766$ & 88.24\% \\
        OGB-Cube delayed & 2.10517 & $0.10688\pm0.01710$ & 94.92\% \\
        \bottomrule
    \end{tabular}
    }
    \caption{Five-block autoregressive latent MSE before and after dense replay adaptation. Adaptation reduces MSE on all eight shifts, with reductions from 45\% to 99\%.}
    \label{tab:prediction_raw}
\end{table}

Matched MSE under the training dynamics is 0.10458 for PushT, 0.11622 for Two-Room, 1.27430 for Reacher, and 0.02131 for OGBench-Cube. These checks use the same task-specific behavior families as the diagnostics under the test-time dynamics.

\subsection{Decoded Predictions}

\begin{figure}[!p]
    \centering
    \includegraphics[width=\textwidth]{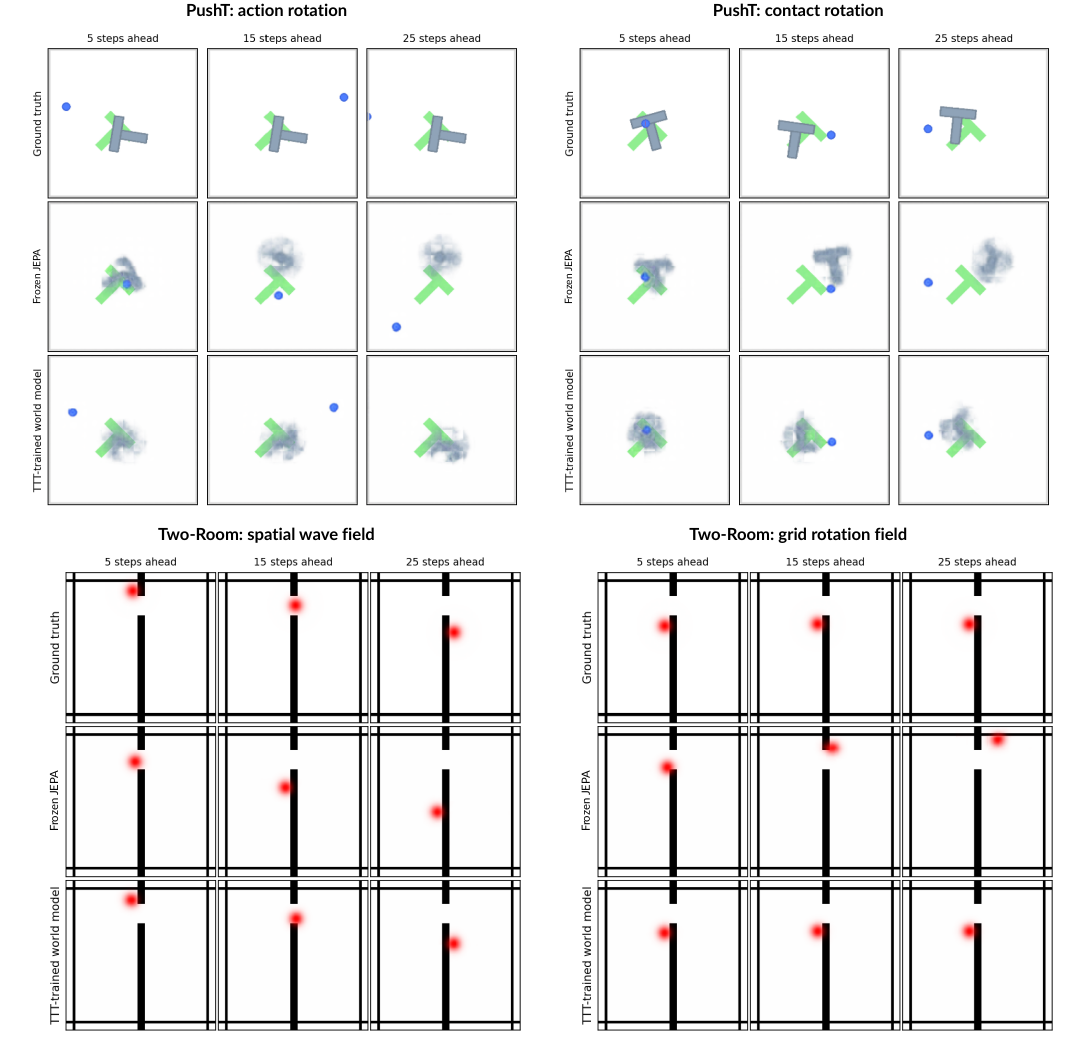}
    \caption{Decoded predictions for both PushT and Two-Room shifts. Rows within each subpanel are ground truth, Frozen JEPA prediction, and TTT-trained-world-model prediction; columns are 5, 15, and 25 raw steps ahead. After adaptation, decoded predictions more closely follow the ground truth across prediction horizons.}
    \label{fig:qualitative_1}
\end{figure}

\begin{figure}[!p]
    \centering
    \includegraphics[width=\textwidth]{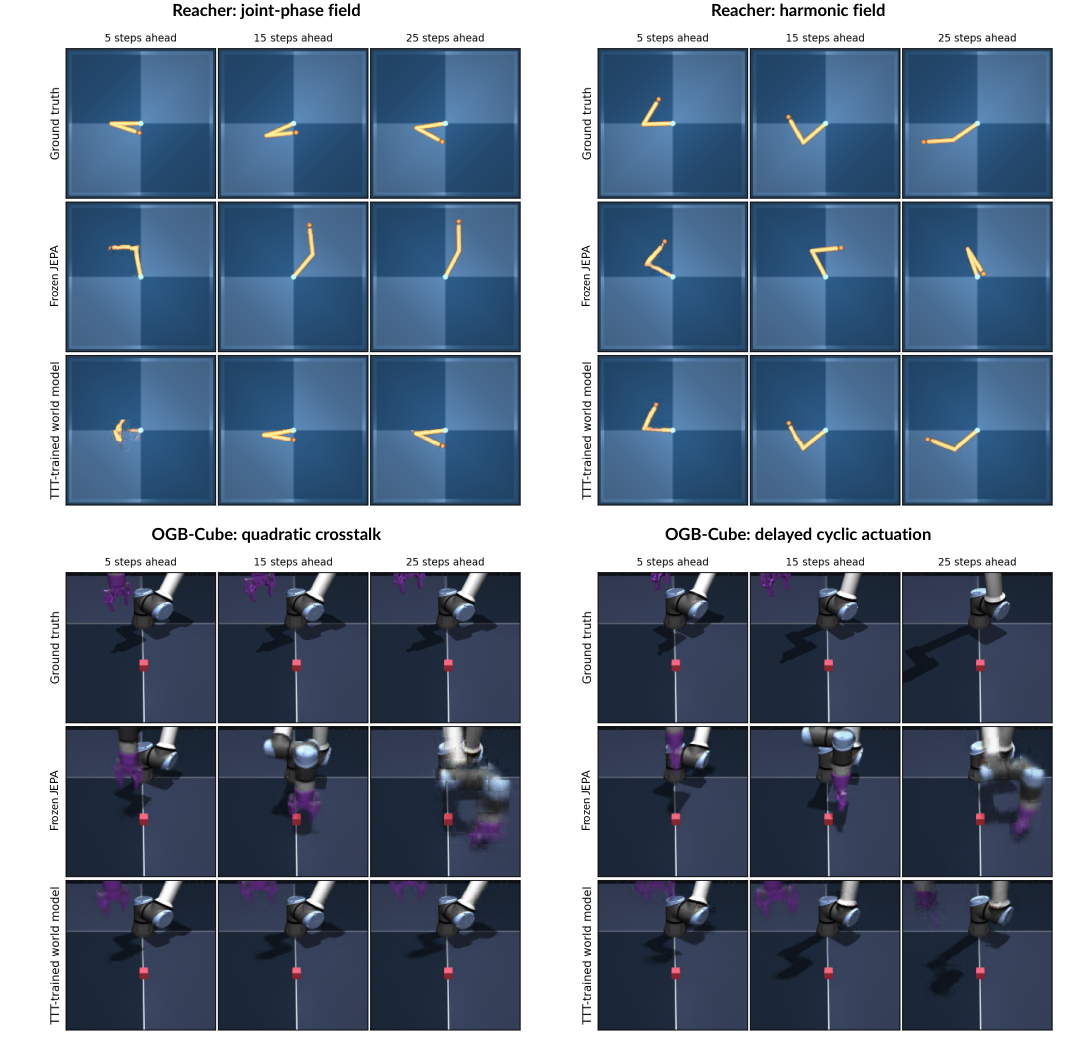}
    \caption{Decoded predictions for both Reacher and OGB-Cube shifts, using the same layout as Figure~\ref{fig:qualitative_1}. After adaptation, decoded predictions more closely follow the ground truth across prediction horizons.}
    \label{fig:qualitative_2}
\end{figure}

Figures~\ref{fig:qualitative_1} and \ref{fig:qualitative_2} decode autoregressive latent predictions with the frozen pretrained decoder. Each panel uses the same initial observation and future action sequence for ground truth, the frozen JEPA world model, and the adapted model. The examples are qualitative diagnostics and are not used to compute planning scores or prediction MSE.

\FloatBarrier

\section{Ablations and Offline References}
\label{sec:appendix_controls}

\subsection{Matched Persistence Control}

Both variants use dense windows from the current episode, batch size 8, the same prediction loss, identical optimizer hyperparameters, the same CEM planner, and the same number of predictor updates. A batch size of 8 is used because many individual episodes do not contain enough windows to form a batch of 64. The persistent variant carries the adapted predictor and optimizer state across episodes. The reset variant restores the pretrained predictor and reinitializes the optimizer at every episode boundary.

\begin{table}[H]
    \centering
    {\small
    \begin{tabular}{llccc}
        \toprule
        Shift & Metric & Persistent & Episodic reset & Difference \\
        \midrule
        \multirow{2}{*}{PushT action rotation} & Best & $0.4683\pm0.0239$ & $0.1049\pm0.0168$ & $+0.3634$ \\
        & AUC & $0.3863\pm0.0247$ & $0.0934\pm0.0016$ & $+0.2929$ \\
        \cmidrule(lr){1-5}
        \multirow{2}{*}{PushT contact rotation} & Best & $0.6393\pm0.0292$ & $0.1735\pm0.0924$ & $+0.4658$ \\
        & AUC & $0.5324\pm0.0352$ & $0.1254\pm0.0127$ & $+0.4069$ \\
        \cmidrule(lr){1-5}
        \multirow{2}{*}{Two-Room spatial wave} & Best & $0.7971\pm0.0016$ & $0.5972\pm0.0135$ & $+0.1999$ \\
        & AUC & $0.7692\pm0.0041$ & $0.5614\pm0.0089$ & $+0.2078$ \\
        \cmidrule(lr){1-5}
        \multirow{2}{*}{Two-Room grid field} & Best & $0.7892\pm0.0157$ & $0.2178\pm0.0061$ & $+0.5714$ \\
        & AUC & $0.7189\pm0.0193$ & $0.2127\pm0.0043$ & $+0.5062$ \\
        \cmidrule(lr){1-5}
        \multirow{2}{*}{Reacher joint phase} & Best & $0.9394\pm0.0717$ & $0.3931\pm0.0275$ & $+0.5463$ \\
        & AUC & $0.7329\pm0.1311$ & $0.3816\pm0.0162$ & $+0.3513$ \\
        \cmidrule(lr){1-5}
        \multirow{2}{*}{Reacher harmonic} & Best & $0.9153\pm0.1150$ & $0.3909\pm0.0379$ & $+0.5244$ \\
        & AUC & $0.7426\pm0.1098$ & $0.3810\pm0.0211$ & $+0.3616$ \\
        \cmidrule(lr){1-5}
        \multirow{2}{*}{OGB-Cube quadratic} & Best & $0.6385\pm0.0085$ & $0.2622\pm0.0013$ & $+0.3763$ \\
        & AUC & $0.5996\pm0.0067$ & $0.2569\pm0.0017$ & $+0.3427$ \\
        \cmidrule(lr){1-5}
        \multirow{2}{*}{OGB-Cube delayed} & Best & $0.6465\pm0.0010$ & $0.1723\pm0.0064$ & $+0.4742$ \\
        & AUC & $0.6119\pm0.0042$ & $0.1686\pm0.0021$ & $+0.4433$ \\
        \midrule
        \multirow{2}{*}{Mean} & Best & 0.7292 & 0.2890 & $+0.4402$ \\
        & AUC & 0.6367 & 0.2726 & $+0.3641$ \\
        \bottomrule
    \end{tabular}
    }
    \caption{Best held-out score and normalized held-out planning-score AUC for the matched persistence-versus-reset control. Entries are means and sample standard deviations over three runs. Persistence improves both metrics on every shift.}
    \label{tab:persistence_control}
\end{table}

\subsection{Continued Training under the Training Dynamics}

We run dense replay for 500 episodes under the unchanged training dynamics as a no-shift control. Across the four tasks, the best score averages 0.679 for the frozen JEPA world model and 0.691 after continued training. Mean AUC is 0.679 for the frozen JEPA world model and 0.676 after continued training. These changes are small and inconsistent across tasks, so continued training under the training dynamics provides no consistent benefit. Table~\ref{tab:base_online_control} reports both metrics for each task.

\begin{table}[H]
    \centering
    {\small
    \begin{tabular}{lccc}
        \toprule
        Task & \makecell{Frozen JEPA\\best/AUC} & \makecell{Continued training\\best} & \makecell{Continued training\\AUC} \\
        \midrule
        PushT & 0.2682 & $0.3599\pm0.0187$ & $0.3247\pm0.0264$ \\
        Two-Room & 0.8039 & $0.8013\pm0.0019$ & $0.7894\pm0.0084$ \\
        Reacher & 0.9995 & $0.9520\pm0.0065$ & $0.9438\pm0.0042$ \\
        OGB-Cube & 0.6439 & $0.6492\pm0.0010$ & $0.6465\pm0.0003$ \\
        \bottomrule
    \end{tabular}
    }
    \caption{Best held-out score and normalized held-out planning-score AUC for dense replay under the unchanged training dynamics. Continued training produces no consistent improvement when the dynamics do not change.}
    \label{tab:base_online_control}
\end{table}

\subsection{Training-Dynamics Checks}

\begin{table}[H]
    \centering
    {\small
    \begin{tabular}{lcccc}
        \toprule
        Task & Planning score & Strict success & Reward-head MSE & Decoder MSE \\
        \midrule
        PushT & 0.2682 & 0\% & $2.54\times10^{-4}$ & $8.10\times10^{-4}$ \\
        Two-Room & 0.8039 & 100\% & $1.24\times10^{-6}$ & $5.21\times10^{-6}$ \\
        Reacher & 0.9995 & 100\% & $1.36\times10^{-6}$ & $3.85\times10^{-5}$ \\
        OGB-Cube & 0.6439 & 100\% & $3.34\times10^{-5}$ & $5.77\times10^{-4}$ \\
        \bottomrule
    \end{tabular}
    }
    \caption{Planning and held-out reward-head and decoder diagnostics under the training dynamics. All reward heads and decoders have low held-out MSE, while PushT is the only task without strict CEM success.}
    \label{tab:base_health}
\end{table}

The epoch-20 upstream validation prediction losses are 0.01291 for PushT, 0.00802 for Two-Room, 0.02702 for Reacher, and 0.01534 for OGBench-Cube. These diagnostics use overlapping random clips, so the episode-held-out errors under the training dynamics in Section~\ref{sec:appendix_prediction} are the stronger generalization check.

PushT's aggregate reward-head MSE is dominated by low-score states. Of 90,300 validation frames, 86,466 have scores below 0.1, while only 114 have scores of at least 0.5. MSE rises to 0.0154 in the score interval $[0.5,0.7)$ and 0.0683 in $[0.7,1.0)$, with systematic underprediction. A score-balanced diagnostic head changes CEM coverage under the training dynamics from 0.2682 to 0.3137 but produces no strict successes with either head, so we retain the original protocol and report this limitation directly.

\subsection{Offline Reference under the Test-Time Dynamics}

We include a data-rich offline reference trained directly under each set of test-time dynamics to contextualize the remaining adaptation gap. To construct it, we collect 3,000 complete episodes from the shifted environment, train a new world model from scratch for 20 epochs, fit a reward head using score labels from the same dynamics, and evaluate on 100 held-out episodes. It is not a deployable baseline and is not a mathematical upper bound because it uses a different data-collection distribution and receives 3,000 complete offline episodes from the shifted environment. Specifically, its collection policy is not task-optimal, whereas \method{} receives on-policy CEM trajectories during test-time training. The persistent result can therefore be slightly higher in some cases. Bracketed intervals below are the mean plus or minus 1.96 standard errors across the 100 episodes.

\begin{table}[H]
    \centering
    {\small
    \begin{tabular}{lcccc}
        \toprule
        Shift & Frozen JEPA & \makecell{\method{}\\best} & \makecell{\method{}\\AUC} & \makecell{Offline model\\\mbox{[95\% CI]}} \\
        \midrule
        PushT action rotation & 0.0804 & $0.490\pm0.022$ & $0.378\pm0.022$ & $0.5667\ [0.5447,0.5886]$ \\
        PushT contact rotation & 0.1268 & $0.619\pm0.033$ & $0.555\pm0.028$ & $0.6321\ [0.6145,0.6497]$ \\
        Two-Room spatial wave & 0.5572 & $0.799\pm0.002$ & $0.728\pm0.012$ & $0.7993\ [0.7970,0.8017]$ \\
        Two-Room grid field & 0.2106 & $0.799\pm0.003$ & $0.725\pm0.019$ & $0.8021\ [0.7993,0.8050]$ \\
        Reacher joint phase & 0.3673 & $0.758\pm0.012$ & $0.619\pm0.019$ & $0.9994\ [0.9993,0.9994]$ \\
        Reacher harmonic & 0.3687 & $0.688\pm0.075$ & $0.518\pm0.012$ & $0.9994\ [0.9994,0.9995]$ \\
        OGB-Cube quadratic & 0.2572 & $0.628\pm0.020$ & $0.530\pm0.009$ & $0.6375\ [0.6322,0.6428]$ \\
        OGB-Cube delayed & 0.1717 & $0.643\pm0.003$ & $0.513\pm0.009$ & $0.6527\ [0.6465,0.6589]$ \\
        \bottomrule
    \end{tabular}
    }
    \caption{Offline references trained under the test-time dynamics. The best held-out score and normalized held-out planning-score AUC follow the definitions in Section~\ref{sec:experimental_setup}. The offline reference is a separately pretrained model evaluated once on 100 held-out episodes. The best \method{} score is within 0.08 of the offline reference on six shifts, with larger gaps on the two Reacher shifts.}
    \label{tab:target_domain_reference}
\end{table}

\FloatBarrier

\begin{figure}[!t]
    \centering
    \includegraphics[width=0.86\textwidth]{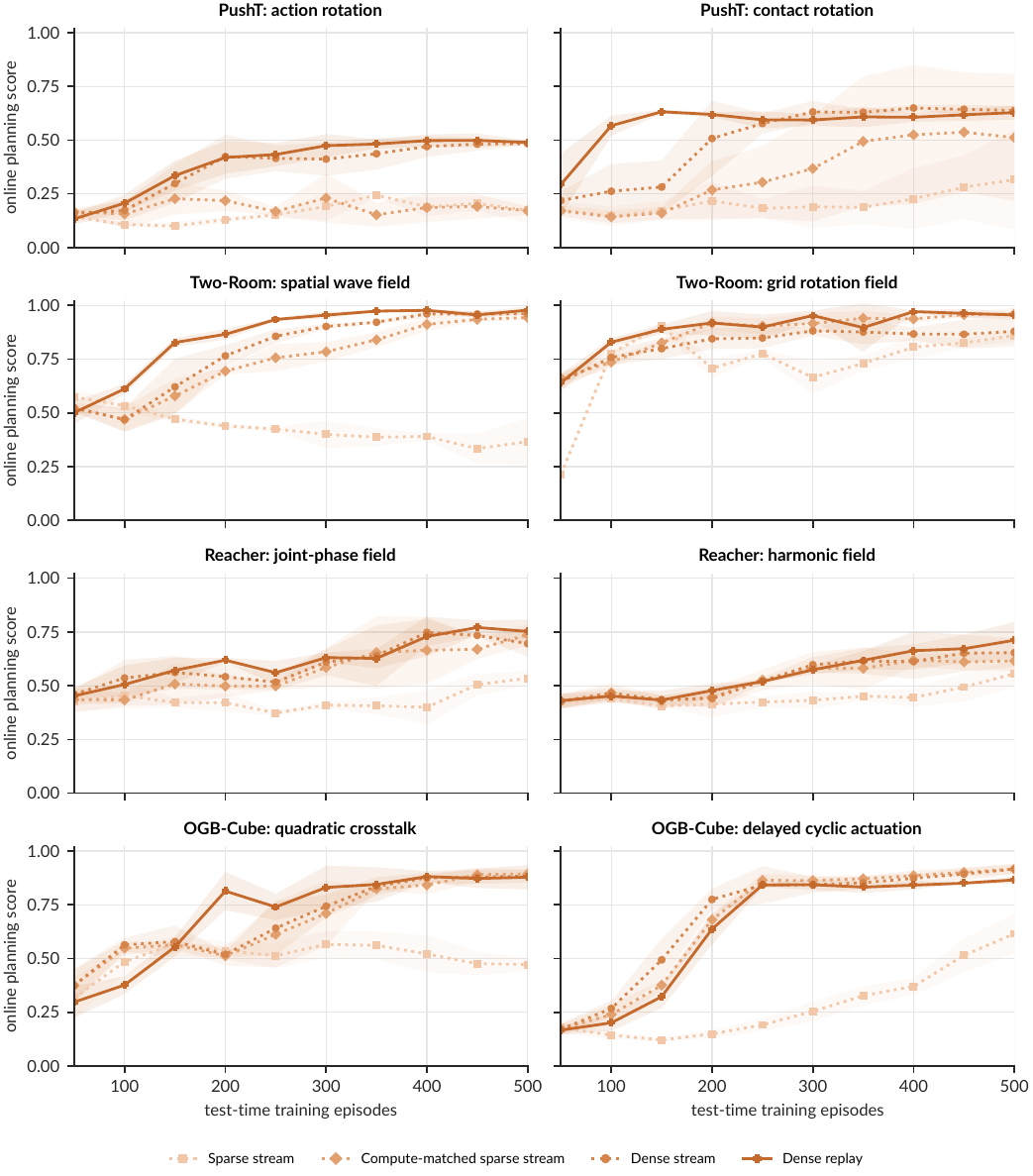}
    \caption{Online test-time performance for \method{} and the three ablations of dense replay. Curves show means $\pm$ sample standard deviations over three runs, binned into consecutive groups of 50 episodes. Dense replay achieves the highest aggregate online score.}
    \label{fig:online_performance}
\end{figure}

\section{Online Test-Time Performance}
\label{sec:appendix_online}

The main paper evaluates planning performance on fixed held-out episodes. Figure~\ref{fig:online_performance} instead reports the scores obtained during the 500 episodes that supply test-time training data. Each point averages a consecutive block of 50 test episodes. These curves include the feedback between the changing planner and the data it collects, so they are noisier than held-out evaluations but show the performance experienced online.